%% file: main.tex
\documentclass[conference]{IEEEtran}
\IEEEoverridecommandlockouts
\usepackage{graphicx,amssymb,amsmath,amsfonts}
\usepackage[ruled,vlined]{algorithm2e}
\usepackage{array}
\usepackage[caption=false,font=normalsize,labelfont=footnotesize,textfont=footnotesize]{subfig}
\usepackage{textcomp}
\usepackage{stfloats}
\usepackage{url}
\usepackage{verbatim}
\usepackage{cite}
\usepackage{xcolor}
\usepackage{siunitx}
\usepackage{tabularx} 
\usepackage{multirow}
\usepackage{balance}
\usepackage{hyperref}
\usepackage{epstopdf}

\usepackage{tikz}
\usepackage{tikz}
\usetikzlibrary{shapes}
\usetikzlibrary{plotmarks}
\usetikzlibrary{shadings}

\begin{document}

\title{Genetic Algorithm-based Routing and Scheduling for Wildfire Suppression using a Team of UAVs\\
}
\author{\IEEEauthorblockN{Josy John}
\IEEEauthorblockA{\textit{Department of Aerospace Engineering} \\
\textit{Indian Institute of Science}\\
Bengaluru, India \\
josyjohn@iisc.ac.in}
\and
\IEEEauthorblockN{Suresh Sundaram}
\IEEEauthorblockA{\textit{Department of Aerospace Engineering} \\
\textit{Indian Institute of Science}\\
Bengaluru, India \\
vssuresh@iisc.ac.in}
}

\maketitle

\begin{abstract}
This paper addresses early wildfire management using a team of UAVs for the mitigation of fires. The early detection and mitigation systems help in alleviating the destruction with reduced resource utilization. A Genetic Algorithm-based Routing and Scheduling with Time constraints (GARST) is proposed to find the shortest schedule route to mitigate the fires as Single UAV Tasks (SUT).  The objective of GARST is to compute the route and schedule of the UAVs so that the UAVS reach the assigned fire locations before the fire becomes a Multi UAV Task (MUT) and completely quench the fire using the extinguisher. The fitness function used for the genetic algorithm is the total quench time for mitigation of total fires. The selection, crossover, mutation operators, and elitist strategies collectively ensure the exploration and exploitation of the solution space,  maintaining genetic diversity, preventing premature convergence, and preserving high-performing individuals for the effective optimization of solutions. The GARST effectively addresses the challenges posed by the NP-complete problem of routing and scheduling for growing tasks with time constraints. The GARST is able to handle infeasible scenarios effectively, contributing to the overall optimization of the wildfire management system.
\end{abstract}


\section{Introduction}
Wildfire scenarios demand rapid mitigation strategies to reduce the destruction of wildlife and wildlands. The wildfires grow in size and number, depending on wind and terrain conditions. Thus, containing the wildfires at an early stage is imperative to curtail the damage. The utilization of multi-UAV systems in wildfire management aims to reduce human interactions in hazardous environments \cite{akhloufi2020unmanned}, \cite{yuan2015survey}. While existing literature has primarily focused on individual or combined \cite{OMS} aspects of search, monitoring, and mitigation, tackling the task allocation dimension in the domain of multi-UAV wildfire management stands out as a critical challenge. Limited works have explored the optimal sequence computation for neutralizing multiple wildfires.  Various multi-UAV multi-task allocation methods have been applied in diverse domains, including search and rescue \cite{Jin2006searchandrescue}, \cite{zhao2016searchandrescue}, job-shop scheduling \cite{Ahn2023jobshop}, vehicle routing problems \cite{Dorling2017routing}, etc. Sequential multi-task allocation proves beneficial when the number of agents is limited and tasks are distributed unevenly across the mission area \cite{sujit2007sequential}. The dynamic nature of fire and limited firefighting resources may lead to uncontrollable fires, making multi-UAV multi-task allocation a challenging research area.

The problem of efficiently scheduling firefighting resources, such as fire engines, air tankers, and helicopters, to contain multiple fires is addressed in \cite{pappis2010scheduling}. A multi-objective hybrid differential evolution particle swarm optimization for the emergency scheduling of fire engines for forest fires is addressed in \cite{Tian2016}. The multi-objective scheduling model is analyzed for the case where the number of fire engines is higher than the fire areas and offers efficient solutions to minimize both the extinguishing time and the number of scheduled fire engines. In \cite{parker2018lazy}, the challenge of assigning heterogeneous agents to tasks with growing costs is tackled using a lazy max-sum algorithm. A model-based multiagent dynamic task assignment for wildfires is discussed in \cite{chen}, aiming to minimize the total completion time. In the above-mentioned works, the number of agents is much higher than the total tasks, and the execution capacity employed is always higher than the optimal value, leading to the wastage of resources. In the context of the Dynamic Multi-Point Dynamic Aggregation (DMPDA) problem, \cite{gao2022mpda} introduces a genetic programming hyperheuristic algorithm. This approach evolves reactive coordination strategies for scenarios where task information is unknown until detection and task demand is time-varying. The solution strategy requires a coalition between agents for the suppression of fires.

The early mitigation of wildfires is essential for minimizing biodiversity loss, ensuring a higher mission success rate, and optimizing resource utilization. The delayed mitigation can lead to fires evolving into Multi-UAV Tasks (MUT), necessitating simultaneous actions and cooperative coalitions among agents. However, forming such coalitions introduces computational complexity and communication overhead. If the fires are contained as a Single UAV Task (SUT), they can be addressed in their early stages with minimal damage, reduced computational demands, and lowered communication overhead. This strategic approach not only enhances the efficiency of wildfire management but also streamlines the coordination among UAVs, ensuring a more effective and resource-efficient mitigation process.

This paper presents a Genetic Algorithm-based Routing and Scheduling with Time constraints (GARST) for wildfire suppression employing a team of UAVs. The primary goal of GARST is to prevent fires from surpassing the quenching capabilities of a single UAV, coupled with the secondary objective of minimizing the total quench time for the fire area. These objectives are aimed at achieving efficient resource allocation and reducing damage to wildlands during wildfire mitigation. The approach is formulated under the assumption that the number of UAVs is significantly less than the total number of fire areas within the mission region. The wildfire mitigation problem is cast as a shortest schedule route problem with time constraints, wherein UAVs must compute routes for the sequential mitigation of fire areas as SUT and minimize the scheduled dynamic quenching times. The NP-complete nature of the problem makes it computationally complex to solve using conventional approaches hence, a routing and scheduling based on Genetic Algorithm (GA) is proposed for the time-sensitive mitigation of wildfires. The effectiveness of the GARST for solving wildfire suppression is evaluated using different numbers of fires. A Monte-Carlo simulation is also conducted with different initial locations of UAV and fire radii. The results show that GARST is successful in evolving a better shortest schedule route to mitigate fires as SUT. The GARST is $100\%$ successful in handling fires up to $4$ times the number of agents and $93\%$ when the number of fires is $5$ times that of the number of UAVs. The main contributions of GARST are summarized as follows:
\begin{enumerate}
\item The wildfire suppression using a homogeneous team of UAVs is formulated as the shortest schedule route problem with time constraints.
\item A genetic algorithm-based solution is proposed for the NP-complete problem of wildfire suppression.
\end{enumerate}

The rest of this paper is organized as follows. Section II provides a review of existing multi-agent task allocation approaches. Section III presents the formulation of the wildfire suppression problem as a shortest schedule route problem. Section IV explains the GARST for wildfire management. Numerical simulation results are given in Section V to verify the performance of the GARST for wildfire suppression, and Section VI concludes the paper.
\section{Related Works}

The wildfire mitigation using a team of UAVs has been framed as a multi-agent multi-task allocation problem in literature. This section briefly summarizes the existing literature on multi-agent multi-task allocation, covering various problem contexts, and discusses its relevance to real-world scenarios in this domain.

\subsection{Multi-task Assignment Problems}
The Consensus-Based Bundle Algorithm (CBBA) for coordinating a fleet of autonomous vehicles addressing multi-assignment problems is discussed in \cite{choi2009CBBA}. CBBA utilizes a decentralized auction-based decision strategy for task selection and employs a consensus routine for conflict resolution, ensuring convergence to conflict-free assignments. Building upon the CBBA, \cite{hunt2014CBGA} explores consensus algorithms for coordinating UAVs in MUT assignments. The Consensus-Based Grouping Algorithm (CBGA) addresses challenges such as MUTs, equipment requirements, and task dependencies. The CBGA enhances UAV cooperation for improved autonomous operations by efficiently reducing communication costs and achieving consensus on MUTs. A Consensus-Based Timetable Algorithm (CBTA) is presented in \cite{wang2022consensus} to address the decentralized simultaneous multi-agent task allocation problem, where multiple agents are needed to perform a task simultaneously. The CBTA minimizes the start time of each task, indirectly aiming to minimize the average start time of all tasks. CBTA performs similarly to the CBBA for single-agent tasks and outperforms the CBGA for multi-agent tasks. A decentralized genetic algorithm approach for multi-agent task allocation to minimize the mission completion time, highlighting its parallelization over agents for efficient use of computational resources is analyzed in \cite{patel2020decentralized}.

\subsection{Multi-task Assignment Problems with Time Constraints}
The challenge of maximizing the number of task allocations in a distributed multi-robot system operating under strict time constraints and fuel limits is addressed in \cite{turner2017distributed}. Performance Impact-MaxAss (PI-MaxAss) follows a two-phase task assignment strategy where a solution generated from an existing PI algorithm \cite{whitbrook2015novel} is iteratively improved using PI-MaxAss to maximize task assignments without repeating the entire allocation procedure. A distributed task allocation problem for maximizing the total number of successfully executed tasks in multi-robot systems considering task deadlines and fuel limits is studied in \cite{wang2023efficient}. The Effective and Efficient Performance Impact (EEPI) algorithm incorporates a novel cost function that minimizes traveling time and prioritizes tasks with earlier deadlines.

\subsection{Multi-task Assignment Problems for Real-world Scenarios}
A heuristic distributed task allocation method for multivehicle multitask assignment problems, particularly in the context of a search and rescue scenario is presented in \cite{zhao2015heuristic}.  The heterogeneous team of UAVs uses a distributed algorithm with local communication to handle rescue missions under various scenarios, such as blocked paths or updated terrain information. The complex problem of multi-robot task allocation, in the context of cooperative tasks in industrial plant inspection, is addressed in \cite{liu2016performance}. It introduces a subpopulation-based genetic algorithm to minimize task completion time with mutation operators and elitism for solving multi-robot task allocation problems. A routing and scheduling problem focusing on the airport ground movement problem is solved using the memetic algorithm for routing in multigraphs with time constraints in \cite{beke2023routing}. The method is tested on real data from two airports, demonstrating its effectiveness in providing solutions within given time budgets.

The existing literature lacks solutions for wildfire scenarios to compute the route and schedule for UAVs for neutralizing tasks with dynamic neutralization time and deadline time. This paper proposes GARST for wildfire suppression using a team of UAVs. 

\section{Problem Definition}
\begin{figure}[!ht]
	\centerline{\includegraphics[width=50mm]{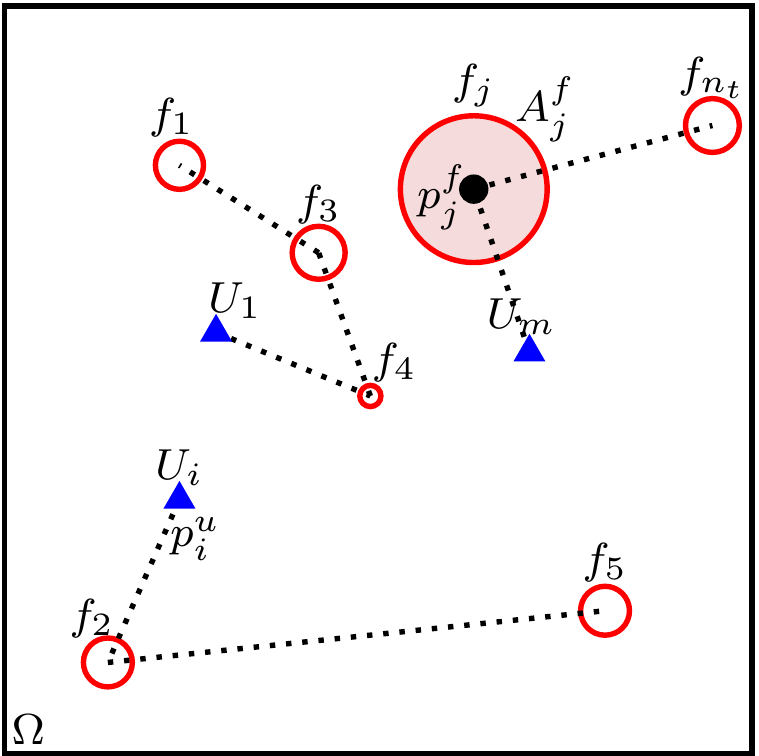}}
	\caption{A wildfire management scenario showing the routing and scheduling for a team of UAVs}
	\label{forestfire_TA}
\end{figure}

A wildfire management scenario is shown in Fig. \ref{forestfire_TA}, where multiple UAVs are deployed in the mission area, $\Omega\subset \mathbb{R}^2$ to mitigate randomly located multiple fires sequentially. The UAVs in the team need to compute the route and schedule for the sequential mitigation of fires to minimize the destruction of wildlands. Each active forest fire area possesses a critical quench rate, surpassing which additional quench rates prove ineffective in reducing the overall quench time \cite{MSCIDC}. This study operates under the assumption that UAVs operate with a quench rate near the critical quench rate. Consequently, introducing additional UAVs does not yield significant performance improvements if the fire scenario remains categorized as a SUT. If the UAVs possess adequate quench capacities to sequentially mitigate fires independently at an early stage of fires, then potential biodiversity loss can be reduced, delays associated with waiting for the coalition UAVs can be avoided, and resource utilization is maximized by sequential mitigation. Therefore, this study addresses the SUT assignment problem, focusing on the mitigation of all fires with a single UAV before they escalate into MUT.

Let $m$ be the number of UAVs, and  $\mathcal{U} \triangleq \{U_1,..., U_m\}$ represent the set of UAVs. The UAVs employed in the mission are assumed to be homogeneous, possessing identical speeds denoted by $v$ in \si{m/s} and quenching capabilities characterized by the area quench rate $\phi^q$. The position coordinate of the $i$-th UAV is represented by $p^u_{i}$. In the context of a typical wildfire scenario, the tasks or fires exhibit dynamic characteristics concerning size and quantity. This dynamic nature also arises from external factors such as wind and terrain conditions, contributing to the dispersion of fire particles, resulting in an increased number of fires characterized by clustered fire spots. Let $n$ denote the number of fires in the mission area, with the assumption that this number is significantly higher than the number of UAVs deployed for the mission, i.e., $n \gg m$. The set of fires is represented as $\mathcal{F} \triangleq \{f_1,..., f_{n}\}$, and the center of the $j\text{th}$ fire spot is located at the position coordinate $p^f_{j}$. Each fire within the cluster is modeled as a point fire with a circular fire profile, adhering to the circular fire model described in \cite{chen}. The initial areas of the fires are heterogeneous, each having different initial areas denoted by $A^f_{j0}$ in \si{m^2}. The fire area grows over time based on the radial spread rate, $\phi^s$ in \si{m/s}. The area and perimeter of the $j\text{th}$ fire point at time, $t$ \si{s} are denoted as $A^f_j(t)$ and $P^f_j(t)$, respectively. The perimeter can be expressed as a function of the fire area, given by $P^f_j(A^f_j) = 2\sqrt{\pi}\sqrt{A^f_j(t)}$. When the $j\text{th}$ fire is under mitigation by a UAV, the rate of change of the fire area can be expressed as a first-order system equation:
\begin{gather}
    \dot{A}^f_j(t)=\phi^sP^f_j(A^f_j)-\phi^q
    \label{firearea}
\end{gather}

The fire area decreases when $\dot{A}^f_j(t)<0$ i.e., $\phi^q>\phi^sP^f_j(A^f_j)$. The critical area, $A^{c}$ is the fire area below which the fire remains as a SUT, and $A^{c}$ can be computed as
\begin{gather}
    A^{c}=\Big(\frac{\phi^q}{2\sqrt{\pi} \phi^s}\Big)^2
\end{gather}
If $A^f_j > A^{c}$, the fire area surpasses the critical area, the task becomes infeasible for a single UAV. The critical time at which the fire area reaches $A^{c}$ serves as the deadline time for the successful mitigation of the $j\text{th}$ fire as a SUT. The deadline time, denoted as $T^{d}_{j}$ for the $j\text{th}$ fire, can be computed as:
\begin{gather}
    T^d_{j}=\frac{\sqrt{A^{c}}-\sqrt{A^f_{j0}}}{\phi^s\sqrt{\pi}}~~  j\in \mathcal{F}
\end{gather}

The overall objective of this work is characterized by the allocation of a sequence of fires to UAVs, ensuring that all UAVs commence fire mitigation along the designated route before the respective deadline times while minimizing the scheduled quench time for the fires along the route of all UAVs. The route, denoted as $\rho_i=\{\rho_{i1},\rho_{i2},...,\rho_{il},...\}$, represents the sequence of tasks executed by $i\text{th}$ UAV. Here, $\rho_{il}$, signifies the task executed by the $i\text{th}$ UAV at the $l\text{th}$ point of route. The $i\text{th}$ UAV is required to arrive at the fire location and initiate the mitigation of the fire area, $\rho_{il}$ before $T^d_{\rho_{il}}$ to constrain the fire area below $A^{c}$ and mitigate the fire as a SUT. The start time of mitigation of $j\text{th}$ fire by $i\text{th}$ UAV is denoted by $T^s_{ij}$ and can be computed as:
\begin{gather}
		T^s_{ij}=
  	\begin{cases}
		\frac{d_{ij}}{v},& l=1\\
		T^c_{ij'}+\frac{d_{j'j}}{v},& l>1
	\end{cases}
\end{gather}
where $d_{ij}$ is the distance from the initial position of $i\text{th}$ UAV to $j\text{th}$ task. $l=1$ indicates that $j$ is the first task in the route of $i\text{th}$ UAV. $j'$ is the previous task assigned to $i\text{th}$ UAV, $T^c_{ij'}$ is the time to complete the previous task, $d_{j'j}$ is the distance from the previous task to the current task.

The constraint for mitigating fires in the route, $\rho_i ~\forall i\in \mathcal{U}$ as SUT can be mathematically defined as:
\begin{gather}
    T^s_{ij}-T^d_{j}<0~~ \forall j\in \rho_{i} 
    \label{SUTcondition}
\end{gather}

The computation of $\rho_{i}$, minimizing the total quench time while satisfying (\ref{SUTcondition}), can further aid in computing the shortest schedule routes. This ensures that all fires are quenched as Single UAV Tasks (SUTs) at an earlier stage of fire development, thereby reducing the loss of biodiversity and maximizing resource utilization. Let $Q^t_{ij}$ be quench time taken by $i\text{th}$ UAV for $j\text{th}$ fire area and $Q^t_{ij}$ can be computed from the solution for Eq. \ref{firearea} assuming fire area at start time, $A^f_j(T^s_{ij})$ as the initial condition and the fire area after quench time as $A^f_j(T^s_{ij}+Q^t_{ij})=0$ \cite{chen}.
\begin{gather}
Q^t_{ij}=\frac{2\phi^q}{(K\phi^s)^2} \text{ln}\Bigg(\frac{\phi^q}{\phi^q-K\phi^sA^f_j(T^s_{ij})}\Bigg)-\frac{2\sqrt{A^f_j(T^s_{ij})}}{K\phi^s}
\end{gather}
where $A^f_j(T^s_{ij})$ is the fire area of $j\text{th}$ fire when $i\text{th}$ agent reaches the fire location and $K=2\sqrt{\pi}$. The quench time for a specific task depends on the start time of the task, which, in turn, is influenced by the completion time of the previous task. Consequently, the quench time for each task is interlinked with the quench times of the previous tasks in the route. The quench time, $T^q_{ij}$ for $i\text{th}$ UAV to mitigate $j\text{th}$ fire area as SUT is defined as
\begin{gather}
		T^q_{ij}= 
            \begin{cases}
              Q^t_{ij} ,& T^s_{ij}< T^d_{j}\\
              \kappa ,& T^s_{ij}\ge T^d_{j}
            \end{cases}
\end{gather}
The quench time is a finite value for all feasible tasks, while it is represented as $\kappa$ for all infeasible tasks. The fire areas that cannot be successfully mitigated as SUT are termed as infeasible tasks. The parameter $\kappa$ is assigned a very high value to impose a penalty on MUTs.

The wildfire mitigation strategies involve routing UAVs to assigned fire locations and scheduling the UAVs for the time required to quench the fire. Consequently, the wildfire management problem can be conceptualized as an integrated routing and scheduling problem. The mitigation of the fire areas before it escalates into a MUT indirectly minimizes resource wastage. The UAVs should mitigate all fire locations while the fire area remains manageable as a SUT for a mission to be considered successful. The objective of the mission is to guarantee a successful mission, minimizing scheduled quench time, and the objective can be mathematically formulated as
\begin{gather}
    \label{perfobj}
    \min_{x_{ij}}~ \sum_{i=1}^{m} \Bigg(\sum_{j=1}^{n} T^q_{ij}(\rho_i)x_{ij}\Bigg)\\
    \label{SUTeq}
    \text{s.t}~ \sum_{i}^{m}x_{ij} \le 1~~\forall j \in \mathcal{F}\\
    \label{const_temp}
    T^s_{ij} < T^d_{j} ~~\forall j \in \rho_i, \forall i\in \mathcal{U}\\
    \label{bin_var}
    x_{ij} \in \{0,1\} ~~\forall (i,j) \in \mathcal{U} \times \mathcal{F}
\end{gather}
The objective in Eq. \ref{perfobj} is formulated as an integer programming problem to compute the optimal route, $\rho_i,~ \forall i\in \mathcal{U}$. The agents can execute multiple tasks sequentially, and the route $\rho_i$ contains the sequence of tasks for $i\text{th}$ agent. Eq. \ref{SUTeq} denotes that each fire area is assigned to a single UAV, and Eq. \ref{const_temp} gives the temporal constraints to quench the fire as SUT. The Eq. \ref{bin_var} indicates the binary decision variable, $x_{ij}=1 ~\forall j\in \rho_i$ and $0$ otherwise. Each UAV can only address one fire at a time, and it must continuously dispense water on the firefront for the scheduled quench time to achieve complete mitigation. After completing the task, the UAV proceeds to the next fire location in the route. 

The wildfire mitigation problem can be considered a derivative of the technician routing and scheduling problem with repair times, belonging to the class of NP-complete problems. In this context, the scheduled quench time serves as the equivalent of repair time in the technician routing problem. The quench time is not constant for a fire due to the dynamic nature of the fire, rendering the problem NP-complete. Hence, the solution of Eq. \ref{perfobj} using conventional optimization approaches will be computationally complex. The GARST is proposed for solving the NP-complete problem of routing and scheduling of wildfire tasks.

\section{Genetic Algorithm-based Routing and Scheduling for Wildfire Management}

The GARST for Wildfire Management is proposed to minimize the destruction of wildlands by efficiently reducing the total quench time and mitigating the fires as SUTs. In the GARST framework, the route information, comprising the sequence of fires assigned to UAVs, is encoded into chromosomes as part of the genetic algorithm. Each chromosome, or individual, represents a potential solution to the problem of finding the shortest schedule route. Initiating the GARST algorithm involves generating a set of chromosomes to form the initial population, with parents selected based on a fitness function that evaluates their effectiveness in minimizing the overall quench time. Through crossover operations, genetic material is exchanged between parents to create offsprings, promoting the exploration of different routes. Additionally, mutation operators are employed to introduce variability within the population, aiding in the discovery of diverse and potentially better solutions. The implementation of an elitist strategy ensures that the best-performing individuals are preserved across generations, preventing the loss of superior solutions. This elitist approach contributes to the stability and efficiency of GARST by retaining the fittest individuals in the population.
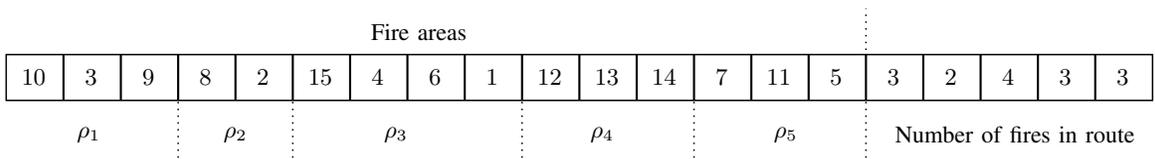
\begin{figure*}[t]
\centering
\input{./TexImages/encoding.tex}
\caption{Representation of an encoded chromosome}
\label{encoding}
\end{figure*}
\subsection{Chromosome Representation}
The path representation, similar to the traveling salesman problem \cite{larranaga1999genetic}, is adopted for chromosome representation of the routing and scheduling problem. The GARST utilizes a path representation with a two-part chromosome structure \cite{carter2006new} for encoding routes. The first part of the chromosome denotes the routes assigned to different UAVs, while the second part signifies the total count of fire areas allocated to each UAV. The chromosome encoding methodology, encompassing a total of $m+n$ genes, is illustrated in Fig. \ref{encoding}. In the first part of the chromosome, the genes represent a permutation of integers ranging from $1$ to $n$, corresponding to the fire areas. Conversely, the second part of the chromosome consists of $m$ genes, indicating the number of fire areas allocated to each respective UAV.

The chromosome represented in Fig. \ref{encoding} shows the routing comprising of $15$ fire areas and $5$ UAVs. The route of UAVs are $\rho_1=\{10, 3, 9\}$, $\rho_2=\{8, 2\}$, $\rho_3=\{15, 4, 6, 1\}$, $\rho_4=\{12, 13, 14\}$, and $\rho_5=\{7, 11, 5\}$. Notably, the values assigned to the genes in the second part are constrained to be $m$ positive integers. The first part of the chromosome does not contain repetitive genes, ensuring uniqueness in the routes assigned to each UAV. Furthermore, the genes in the second part are subject to the constraint that their sum equals the total number of fires to be mitigated, thereby ensuring a valid chromosome representation. Even though the chromosomes may meet the validity criteria, not every valid chromosome guarantees the successful mitigation of all fires as SUTs.

\begin{algorithm}[htbp]
\DontPrintSemicolon
	\caption{GARST Algorithm}\label{GARSTAlg}
        \KwData {Input $m$,~$n$,~$N_g$,~ $\mathbb{G}^0$ with $N^0_c$ chromosomes}
        \KwResult {$\mathbb{G}^{N_g}$,~$J_{best}$}
		\While{$k<N_g$}
		{\textup{Evaluate} fitness, $J$ $\forall g^k_c\in\mathbb{G}^k$ \;
        $\mathbb{G}^k_{best}$=Selection($\mathbb{G}^{k}$):~\textup{Best chromosomes as parents}\;
        $\mathbb{G}^{k+1}$= Crossover($\mathbb{G}^k_{best}$):~\textup{Offsprings}\;
        \For{$c=1 ~\textup{to}~ N^{k+1}_c$}{
        \If{$rand<\mu ~\textup{or}~I^k_c>\gamma$}{
        $g^{k+1}_c$=Mutation($g^{k+1}_c$)\;}}
        $\mathbb{G}^{k+1}$=Elitist($\mathbb{G}^k$)~$\cup$~$\mathbb{G}^{k+1}$:~\textup{New generation}\;
        $k=k+1$
        }
        \textup{For all individuals in} $\mathbb{G}^{N_g}$\textup{evaluate} $J$\;
        $J_{best}=min(J^{N_g})$\;
        Chromosome with $J_{best}$ is the route schedule solution
\end{algorithm}
\subsection{Population Initialization}
The initiation of the population in GA is a crucial step influencing the efficiency of the algorithm in exploring the solution space. The initial population is denoted by set, $\mathbb{G}^0$. In GARST, the challenge is heightened by the constraint of starting fire mitigation before the deadline time. This constraint narrows the solution space, and relying solely on random initialization may yield infeasible routes, hindering GA convergence. To address this, GARST adopts a fitness-based initialization, allowing a maximum of four infeasible tasks. This strikes a balance between exploration and feasibility, promoting a diverse yet feasible initial population. As the number of fires increases, effective initialization becomes more significant. In such scenarios, seeding the initial population with near-optimal solutions becomes crucial for navigating the complex solution space, enhancing convergence toward optimal or near-optimal solutions \cite{beke2021comparison}. The population initialization strategies in GARST contribute to the algorithm's robustness and its potential to generate high-quality solutions in subsequent generations.

\subsection{Fitness Function}
The fitness function is a fundamental component of GA that evaluates the suitability of individual solutions within the population. In GARST, the fitness function, $J$ is formulated as the total quench time associated with the route encoded in the chromosome. 
\begin{gather}
    \label{fitness}
    J= \sum_{i=1}^{m} \Bigg(\sum_{j=1}^{n} T^q_{ij}\Bigg) ~~\forall j \in \rho_i, ~\forall i\in \mathcal{U}
\end{gather}
The fitness function for GARST integrates key parameters, including the scheduled quench time for fires, the routes assigned to UAVs, and the route length of each UAV. This close relation with the route generation process within GARST aligns the fitness evaluation of each chromosome to the overall mission objectives. An essential feature of the fitness function involves the imposition of penalties on $T^q_{ij}$, resulting in elevated fitness values corresponding to the number of infeasible tasks. The penalty-enabled fitness function guides the GARST toward solutions that not only minimize quench time but also adhere to constraints and objectives related to route optimization.

\subsection{Genetic Operators}

Genetic operators are fundamental components of GA-based methodologies, playing a crucial role in steering the population evolution towards optimal or near-optimal solutions across successive generations. GARST relies on three primary genetic operators: selection, crossover, and mutation.

\subsubsection{Selection}The selection operator determines the individuals chosen as parents based on their fitness, favoring those with superior traits. The set of chromosomes in the $k$th generation is denoted by $\mathbb{G}^k$ and $c$th chromosome in $k$th generation is denoted as $g^k_c$. The total number of chromosomes in the population of $k$th generation is denoted as $N^k_c$. The set of best-selected individuals in the $k$th generation is denoted as $\mathbb{G}^k_{best}$. Individuals with shorter scheduled quench times and a higher degree of adherence to constraints are more likely to be selected as parents during the selection process. This promotes the propagation of favorable traits in the population across subsequent generations.

\subsubsection{Crossover}The crossover operator in GARST combines genetic information from selected parents to generate new offspring with diverse characteristics. A single-point crossover with a repair mechanism is employed for the generation of valid offspring from parents. The crossover point is chosen randomly, and the single-point crossover helps in maintaining some subroutes. This process ensures that the resulting routes in the offspring maintain the order and diversity of tasks from both parents, contributing to the exploration of the solution space.

\subsubsection{Mutation}The mutation operator in GARST introduces random changes to individual chromosomes, promoting genetic diversity within the population. This mutation process is contingent upon a mutation probability, $\mu$, and the infeasible task ratio of the chromosome. The infeasible task ratio of a chromosome represents the proportion of infeasible tasks to the total tasks and the infeasibility task ratio for $g^k_c$ chromosome is denoted as $I^k_c$. All chromosomes with an infeasible task ratio greater than a threshold value, $\gamma$ is mutated. The infeasibility-based mutation helps retain offspring with lower infeasible tasks and mutate offspring with higher infeasible tasks. Specifically, a swap mutation operator is applied separately to two parts of the chromosome, targeting tasks with the highest probability of being infeasible. The threshold value on the infeasibility-based mutation plays a role in maintaining diverse individuals in future generations. This mutation strategy aims to ensure genetic diversity in the population, mitigating the risk of premature convergence to suboptimal solutions. 

\begin{figure}[h]
\centering
\subfloat[{} \label{routing15f5a}]{\includegraphics[trim=10 2 5 2,clip,width=7cm]{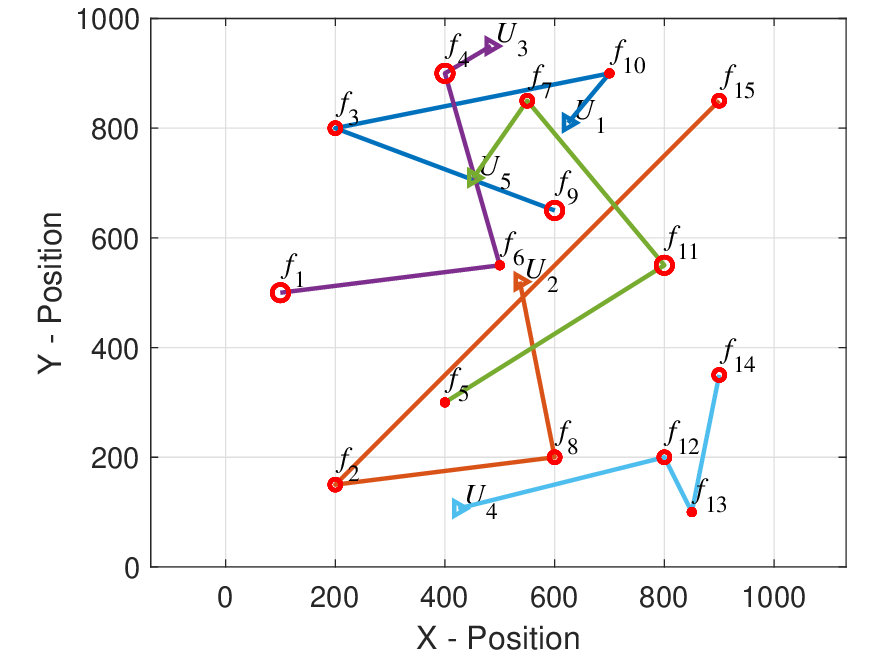}}\\
\subfloat[{} \label{routing20f5a}]{\includegraphics[trim=10 2 5 2,clip,width=7cm]{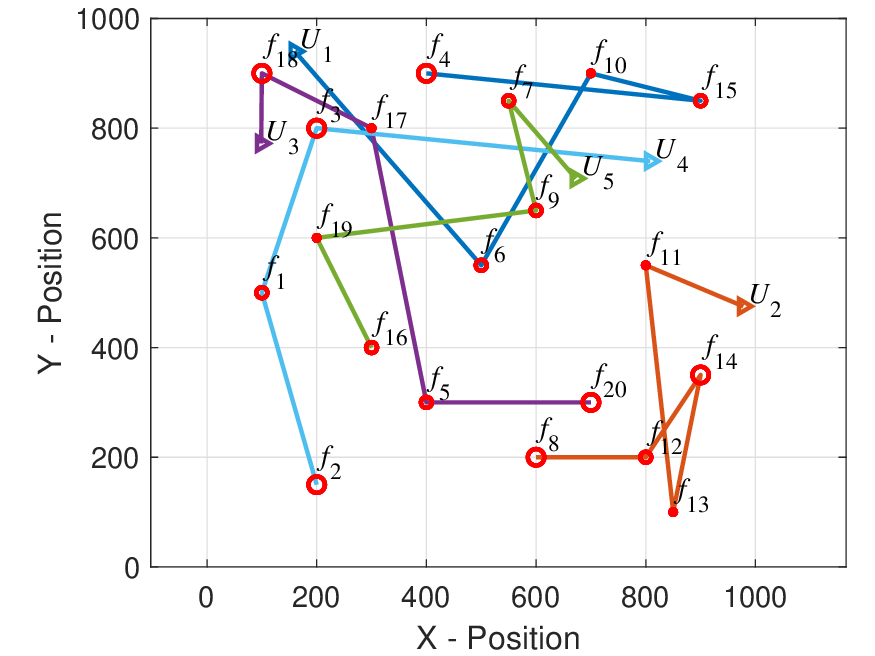}}\\
\subfloat[{} \label{routing25f5a}] {\includegraphics[trim=10 2 5 2,clip,width=7cm]{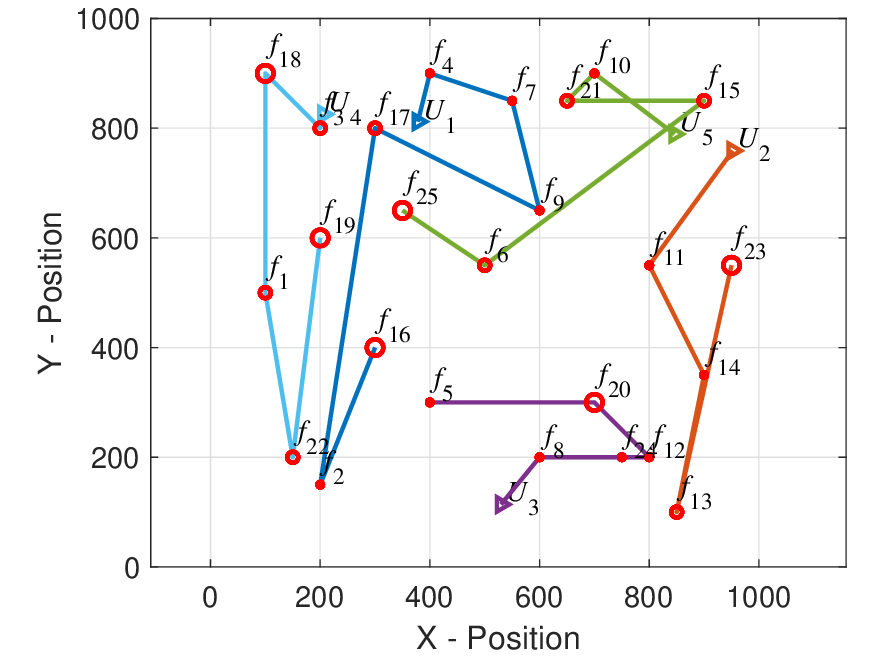}}
\caption{Routing and scheduling solution for a scenario with 5 UAVs (a) 15 fires (b) 20 fires, and (c) 25 fires.}
\label{routingseq}
\end{figure}

\subsubsection{Elitism}The elitist strategy is implemented during offspring generation, where parents with the best fitness values are retained without undergoing any genetic operations, ensuring their continuation into the next generation. This ensures that the top-performing individuals persist across generations, preventing the loss of beneficial genetic material. The next generation of individuals will be the union of elitist parents and the set of unique offspring generated from parents after crossover and mutation. The orchestration of these genetic operators in GARST is carefully designed to strike a balance between exploration and exploitation, promoting the discovery of high-quality solutions for efficient wildfire management. The operational workflow of GARST, detailing the genetic algorithm processes, is depicted in Algorithm \ref{GARSTAlg}.

\section{Results and Discussion}
The GARST is assessed within a $1$ km x $1$ km area, considering varying numbers of fire locations for $5$ UAVs. All fire locations possess a circular profile, with initial fire radii randomly chosen between $5$ to $15$ \si{m}. The UAVs are initialized from diverse locations within the search area and assumed to travel at a constant speed of $20$ m/s. In this study, different UAVs are assumed to fly at different altitudes. Each UAV has a quench capability of $20$  \si{m^2/s}. The simulations are performed in MATLAB R$2022$b environment with an Intel Core-i$7$, $3.2-$GHz processor, and $16-$GB memory.

The initial population size for the GARST is set to $10$, and the population evolves through $50$ generations. A fixed crossover probability of $0.8$ and a mutation probability of $0.01$ are chosen for the genetic operations.  The population size is permitted to vary across generations, while an elitist count of $5$ ensures the retention of superior individuals in each generation. This setup enables GARST to navigate the complex solution space efficiently, seeking optimal or near-optimal solutions for wildfire management in diverse scenarios.

The GARST is evaluated for different numbers of fires, and the routing scheduling solutions for sample scenarios with $15$, $20$, and $25$ fires are shown in Fig. \ref{routingseq}. The representation of the routing schedule illustrates that opting for a greedy solution, where each UAV addresses the nearest small fires, may not result in a successful mission. This challenge is particularly pronounced as the scenarios involve the neutralization of dynamic tasks with time constraints. The fire may escalate into a MUT in wildfire suppression, resulting in infeasible task scenarios. To overcome this, route selection should be strategic, ensuring the mitigation of all fires as SUT with the shortest schedule. This strategic approach may involve choosing distant and larger fires based on their deadline times, thereby enhancing the overall effectiveness of the wildfire management mission.

The effectiveness of GARST for a wildfire scenario can be assessed through the fitness function values in both the initial and final populations. The key performance indices, such as the fitness of the best individual in the final population ($J^{N_g}_{best}$), the average fitness of individuals in the final population ($J^{N_g}_{avg}$), the fitness of the best individual in the initial population ($J^0_{best}$), and the average fitness of individuals in the initial population ($J^0_{avg}$), serve as benchmarks for evaluation. The performance indices for sample scenarios with different numbers of fires are presented in Table \ref{GARSTperf}. These metrics provide insights into the evolution of better individuals over generations and the capability to generate more feasible routes.

\begin{table}[htbp]
\caption{Performance of GARST for different numbers of fires}
\centering
\begin{tabular}{|c|c|c|c|c|}
\hline
$n$          & $J^0_{best}$ & $J^0_{avg}$ & $J^{N_g}_{best}$ & $J^{N_g}_{avg}$ \\ \hline
$15$ & $16.02$      & $29.93$     & $12.68$    & $13.62$   \\ \hline
$20$ & $45.25$      & $64.17$     & $30.37$    & $31.08$   \\ \hline
$25$ & $27.45$      & $69.02$     & $26.76$    & $34.60$   \\ \hline
\end{tabular}
\label{GARSTperf}
\end{table}

The average fitness value of the final population has been decreased to more than $50\%$ of the initial population in all the cases. The substantial decrease in the average fitness value from the initial to the final population highlights the ability of GARST to explore and identify feasible assignments. The ability of the algorithm to navigate a vast search space and converge to feasible solutions is crucial, given the challenging nature of sequential allocation for growing tasks with time constraints. For scenarios involving $25$ fires, seeding the initial population strategically with a near-optimal solution was necessary for convergence.  This emphasizes the pivotal role of the initial population in achieving successful results, especially when the task-to-agent ratio exceeds $4$. Even though the reduction in fitness of the best individual is smaller in the case of $n=25, m=5$,  the evolution from an initial population with only $3$ feasible routes to a final population with $54$ feasible routes is noteworthy.

\begin{table}[htbp]
\caption{Average Performance of GARST for Monte-Carlo analysis }
\centering
\begin{tabular}{|c|c|c|c|c|}
\hline
$n$  & \begin{tabular}[c]{@{}c@{}}Success\\ rate($\%$)\end{tabular} & \begin{tabular}[c]{@{}c@{}}Mean\\ completion \\ time (min)\end{tabular} & \begin{tabular}[c]{@{}c@{}}Mean \\ quench \\ time (min)\end{tabular} & \begin{tabular}[c]{@{}c@{}}Mean\\ FER\end{tabular} \\ \hline
$15$ & $100$                                                  & $5.18 $                                                                 & $13.75 $                                                             & $1.10$                                            \\ \hline
$20$ & $100$                                                  & $9.86$                                                                  & $29.03 $                                                             & $1.90$                                            \\ \hline
$25$ & $93$                                                   & $15.23 $                                                                & $42.88$                                                              & $1.95$                                            \\ \hline
\end{tabular}
\label{MCperf}
\end{table}

The GARST performance is further evaluated with Monte-Carlo analysis for $100$ iterations by varying the initial positions of UAVs and radii of fires for the same set of fire locations. The performance indices used to evaluate the average performance are success rate, mean completion time, mean quench time, and mean Fire Expansion Ratio (FER) for the $100$ iterations of the Monte-Carlo analysis. The success rate is the percentage of successful iterations among the total number of iterations in the Monte-Carlo analysis. The FER is defined as the ratio of an increase in the area of the fire to the initial area of the fire at the start of the mission \cite{MSCIDC}. The mean completion time is the average value of the time required to mitigate all the fires completely. The mean quench time is the average of the total quench time required to mitigate the fires.

The average performance indices of the GARST from the Monte-Carlo analysis are presented in Table \ref{MCperf}. The mean completion time, mean quench time, and mean FER increase with the increase in the number of fires. The lower value of FER indicates that the UAVs reached the fire location faster, preventing further destruction of wildlands. The GARST is $100\%$ successful in quenching all the fire areas as SUTs in the cases of $15$ and $20$ fires. The GARST achieved a success rate of $93\%$ for the $25$ fire scenario. The search space increases with an increase in tasks, and the constraints on time become more prominent with the increased overall fire area diminishing the success rate of finding optimal solutions in larger search spaces. The sequential allocation for growing tasks with time constraints depends on the initial area of fire, and the failure cases had more fires with a larger initial area. This also points out the importance of early wildfire management systems to search, detect, and mitigate fires at early stages.

\section{Conclusions}
This paper presents GARST for early wildfire suppression using a team of UAVs. The method generates the shortest schedule routes for the UAVs for the mitigation of multiple fire areas with time constraints for SUT mitigations. The strategy of mitigating fires as SUTs, combined with the objective of minimizing total quench time through the fitness function, effectively reduces the overall destroyed area and guarantees early suppression of wildfires. The parent selection and elitism mechanism helps in retaining the best individuals in future generations. The proposed crossover and mutation operators balance the search between feasible and infeasible assignments to help generate a better population for future generations. The GARST effectively handled wildfire scenarios with time constraints by evolving superior successful solutions with reduced total quench time. The GARST has been analyzed with Monte-Carlo simulations with different numbers of fire areas. The GARST has a $100\%$ success rate for the number of fires up to $4$ times the number of UAVs and a $93\%$ for $5$ times the number of UAVs. The sequential mitigation of fire areas as SUTs, while minimizing the total quench time, reduces the total burnt wildland with reduced resources.

The future work includes the development of a scalable decentralized approach with better genetic operators for the time-sensitive growing tasks to improve the quality of the final population and convergence of the algorithm.
\bibliographystyle{IEEEtran}
\balance
\bibliography{forestfire_SRT}
\end{document}

%% file: TexImages/encoding.tex
\begin{tikzpicture}[x=1.2in,y=1.2in, scale=1]
\node[inner sep=0.5ex] at (1.8,0.3) {\small $\text{Fire areas}$};
\draw[line width=0.2mm, color=black] (0.0,0.0) rectangle (5,0.2);
\node[inner sep=0.5ex] at (0.12,0.1) {\small $10$};
\draw[line width=0.2mm,thick] (0.25,0.0) -- (0.25,0.2);
\node[inner sep=0.5ex] at (0.37,0.1) {\small $3$};
\draw[line width=0.2mm,thick] (0.5,0.0) -- (0.5,0.2);
\node[inner sep=0.5ex] at (0.62,0.1) {\small $9$};

\node[inner sep=0.5ex] at (0.36,-0.15) {\small $\rho_1$};
\draw[line width=0.2mm,dotted] (0.75,-0.25) -- (0.75,0.0);

\draw[line width=0.2mm,thick] (0.75,0.0) -- (0.75,0.2);
\node[inner sep=0.5ex] at (0.87,0.1) {\small $8$};
\draw[line width=0.2mm,thick] (1.0,0.0) -- (1.0,0.2);
\node[inner sep=0.5ex] at (1.12,0.1) {\small $2$};

\node[inner sep=0.5ex] at (1,-0.15) {\small $\rho_2$};
\draw[line width=0.2mm,dotted] (1.25,-0.25) -- (1.25,0.0);

\draw[line width=0.2mm,thick] (1.25,0.0) -- (1.25,0.2);
\node[inner sep=0.5ex] at (1.37,0.1) {\small $15$};
\draw[line width=0.2mm,thick] (1.5,0.0) -- (1.5,0.2);
\node[inner sep=0.5ex] at (1.62,0.1) {\small $4$};
\draw[line width=0.2mm,thick] (1.75,0.0) -- (1.75,0.2);
\node[inner sep=0.5ex] at (1.87,0.1) {\small $6$};
\draw[line width=0.2mm,thick] (2.0,0.0) -- (2.0,0.2);
\node[inner sep=0.5ex] at (2.12,0.1) {\small $1$};

\node[inner sep=0.5ex] at (1.7,-0.15) {\small $\rho_3$};
\draw[line width=0.2mm,dotted] (2.25,-0.25) -- (2.25,0.0);

\draw[line width=0.2mm,thick] (2.25,0.0) -- (2.25,0.2);
\node[inner sep=0.5ex] at (2.37,0.1) {\small $12$};
\draw[line width=0.2mm,thick] (2.5,0.0) -- (2.5,0.2);
\node[inner sep=0.5ex] at (2.62,0.1) {\small $13$};
\draw[line width=0.2mm,thick] (2.75,0.0) -- (2.75,0.2);
\node[inner sep=0.5ex] at (2.87,0.1) {\small $14$};

\node[inner sep=0.5ex] at (2.6,-0.15) {\small $\rho_4$};
\draw[line width=0.2mm,dotted] (3.0,-0.25) -- (3.0,0.0);

\draw[line width=0.2mm,thick] (3.0,0.0) -- (3.0,0.2);
\node[inner sep=0.5ex] at (3.12,0.1) {\small $7$};
\draw[line width=0.2mm,thick] (3.25,0.0) -- (3.25,0.2);
\node[inner sep=0.5ex] at (3.37,0.1) {\small $11$};
\draw[line width=0.2mm,thick] (3.5,0.0) -- (3.5,0.2);
\node[inner sep=0.5ex] at (3.62,0.1) {\small $5$};

\draw[line width=0.2mm,dotted] (3.75,0.22) -- (3.75,0.4);
\node[inner sep=0.5ex] at (3.4,-0.15) {\small $\rho_5$};
\draw[line width=0.2mm,dotted] (3.75,-0.25) -- (3.75,0.0);

\draw[line width=0.2mm,thick] (3.75,0.0) -- (3.75,0.2);
\node[inner sep=0.5ex] at (3.87,0.1) {\small $3$};
\draw[line width=0.2mm,thick] (4.0,0.0) -- (4.0,0.2);
\node[inner sep=0.5ex] at (4.12,0.1) {\small $2$};
\draw[line width=0.2mm,thick] (4.25,0.0) -- (4.25,0.2);
\node[inner sep=0.5ex] at (4.37,0.1) {\small $4$};
\draw[line width=0.2mm,thick] (4.5,0.0) -- (4.5,0.2);
\node[inner sep=0.5ex] at (4.62,0.1) {\small $3$};
\draw[line width=0.2mm,thick] (4.75,0.0) -- (4.75,0.2);
\node[inner sep=0.5ex] at (4.87,0.1) {\small $3$};

\node[inner sep=0.5ex] at (4.4,-0.15) {\small $\text{Number of fires in route}$};

\end{tikzpicture}